\documentclass{article}
\pdfoutput=1

\usepackage{aaai}
\usepackage{times}
\usepackage{helvet}
\usepackage{courier}
\usepackage{graphicx}
\usepackage[table]{xcolor}
\usepackage{setspace} 
\usepackage{amsmath}

\newcommand{\rotunde}{\textsc{Rotunde}\xspace}
\newcommand{\rotundeTxt}{{Rotunde}\xspace}

\usepackage[ruled,vlined]{algorithm2e}

\usepackage{array}
\usepackage{multirow}

\newcommand\MyBox[3]{
  \fcolorbox[gray]{0}{#1}{\lower0.1cm
    \vbox to 0.55cm{\vfil
      \hbox to 0.55cm{\hfil {\color{#2}#3}\hfil}
      \vfil}%
  }%
}

\newcommand{\beqn}{\begin{eqnarray}}
\newcommand{\eeqn}{\end{eqnarray}}
\newcommand{\sbeqn}{\[\begin{array}{ll}}
\newcommand{\seeqn}{\end{array}\]}

\begin{document}

\title{ \rotunde\ --- A Smart Meeting Cinematography Initiative\\{\small Tools, Datasets, and Benchmarks for Cognitive Interpretation and Control}}

\author{Mehul Bhatt \and Jakob Suchan \and Christian Freksa \\Spatial Cognition Research Center (SFB/TR 8)\\University of Bremen, Germany}

\maketitle
\bibliographystyle{aaai}

\medskip
\medskip

\section{Smart Meeting Cinematography}
We construe \emph{smart meeting cinematography} with a focus on professional situations such as meetings and seminars, possibly conducted in a distributed manner across socio-spatially separated groups.

The basic objective in smart meeting cinematography is to interpret professional interactions involving people, and automatically produce dynamic recordings of discussions, debates, presentations etc in the presence of multiple communication modalities. Typical modalities include gestures (e.g., raising one's hand for a question, applause), voice and interruption, electronic apparatus (e.g., pressing a button), movement (e.g., standing-up, moving around) etc.

\textbf{The Rotunde Initiative}.\quad Within the auspices of the smart meeting cinematography concept, the preliminary focus of the \rotundeTxt\ initiative concerns scientific objectives and outcomes in the context of the following tasks:

\begin{itemize}

\item people, artefact, and interaction tracking

\item human gesture identification and learning, possibly closed under a context-specific taxonomy 

\item high-level cognitive interpretation by perceptual narrativisation and commonsense reasoning about \emph{space, events, actions, change, and interaction}

\item real-time dynamic collaborative co-ordination and self-control of sensing and actuating devices such as pan-tilt-zoom (PTZ) cameras in a \emph{sense-interpret-plan-act} loop

\end{itemize}

Core capabilities that are being considered involve recording and semantically annotating individual and group activity during meetings and seminars from the viewpoint of:

\begin{itemize}

\item computational narrativisation from the viewpoint of declarative model generation, and semantic summarisation

\item promotional video generation

\item story-book format digital media creation

\end{itemize}

These capabilities also directly translate to other applications such as security and well-being (e.g., people falling down) in public space (e.g., train-tracks) or  other special-interest environments (e.g., assisted living in smart homes).

\begin{figure*}[t]
  \centering
    \begin{tabular}{ c c c c} 
      \includegraphics[height=0.17\textwidth]{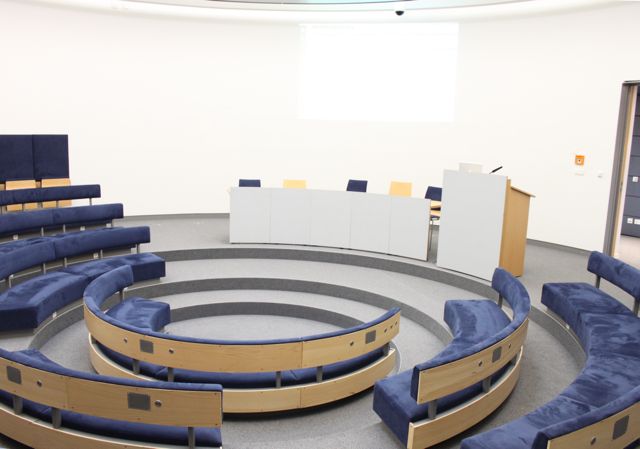} &
      \includegraphics[height=0.17\textwidth]{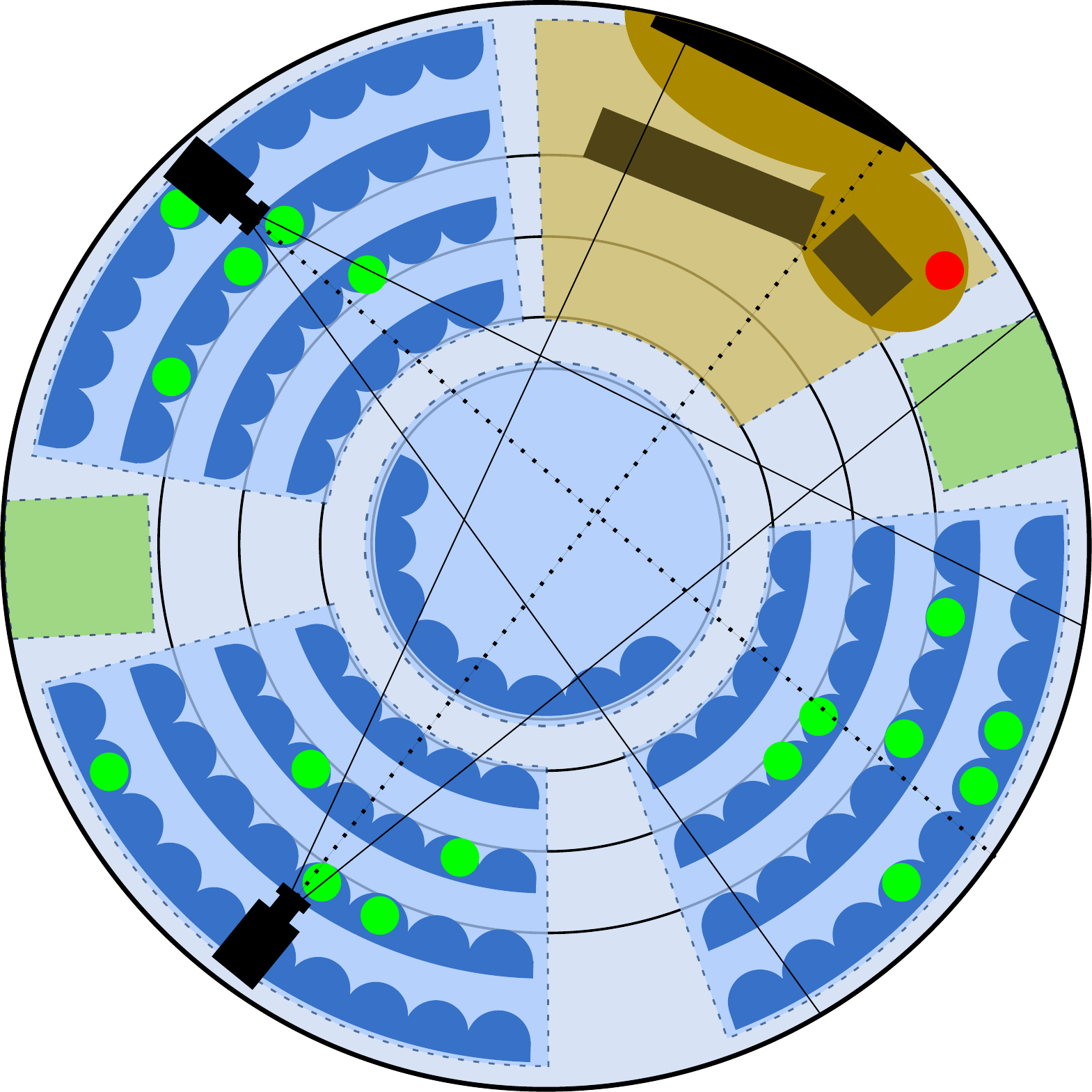} &
      \includegraphics[height=0.17\textwidth]{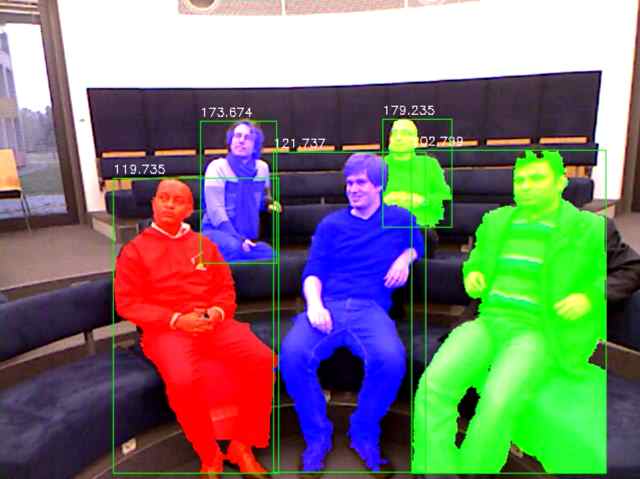} &
      \includegraphics[height=0.17\textwidth]{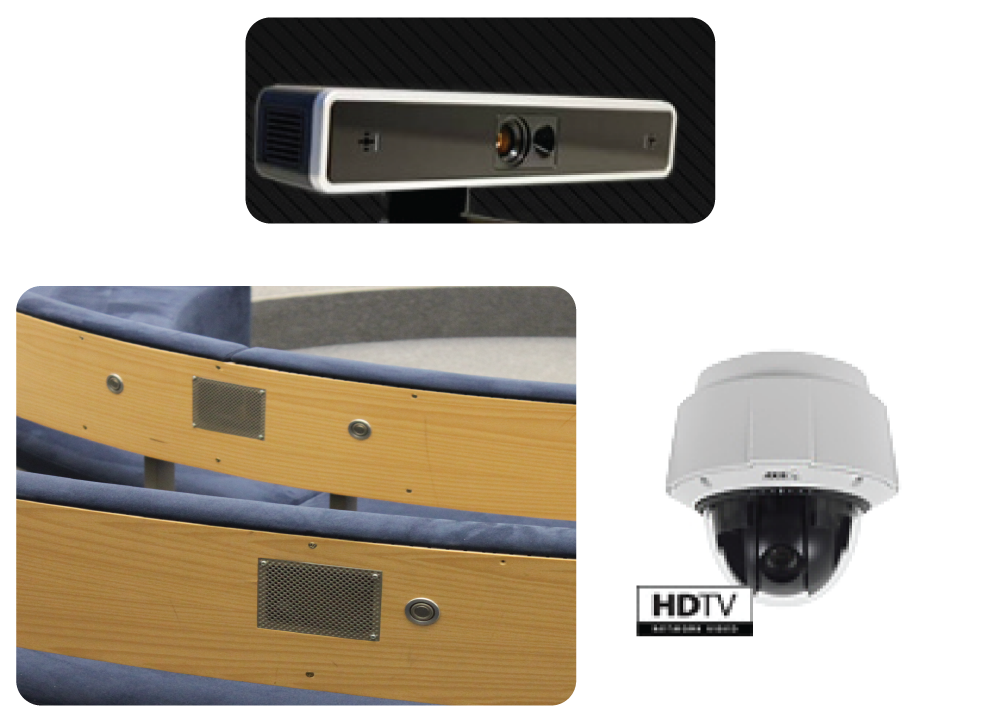} \\
      (a) The Meeting Room & (b) Scenario scheme & (c) People tracking & (d) Sensors \\
    \end{tabular}
  \caption{The Rotunde Setup}
  \label{fig:rotunde-scenario}
\end{figure*}

An example setup for Rotunde is illustrated in Fig. \ref{fig:rotunde-scenario}; this represents one instance of the overall situational and infrastructural setup for the smart meeting cinematography concept.

\section{Cognitive Interpretation of Activities:\\General Tools and Benchmarks}
From the viewpoint of applications, the long-term objectives for the \rotundeTxt\ initiative are to develop benchmarks and general-purpose tools (A--B):

\subsubsection*{A. Benchmarks}
Develop functionality-driven benchmarks with respect to the interpretation and control capabilities of human-cinematographers, real-time video editors, surveillance personnel, and typical human performance in everyday situations

\subsubsection*{B. Tools}
Develop general tools for the commonsense cognitive interpretation of dynamic scenes from the viewpoint of visuo-spatial cognition centred perceptual narrativisation \cite{Bhatt2013-CMN}. 

Particular emphasis is placed on declarative representations and interfacing mechanisms that seamlessly integrate within large-scale cognitive (interaction) systems and companion technologies consisting of diverse AI sub-components. For instance, the envisaged tools would provide general capabilities for high-level commonsense reasoning about space, events, actions, change, and interaction encompassing methods such as \cite{Bhatt:RSAC:2012}:

\begin{itemize}

	\item  \emph{geometric and spatial reasoning} with constraint logic programming \cite{bhatt-et-al-2011}

	\item \emph{integrated inductive-aductive reasoning} \cite{dubba-bhatt-2012} with inductive and abductive logic programming
	
		\item \emph{narrative-based postdiction} (for detecting abnormalities) with answer-set programming \cite{CR-2013-Narra-CogRob}
		
	\item \emph{spatio-temporal abduction}, and \emph{high-level control and planning} with action calculi such as the event calculus and the situation calculus respectively \cite{Bhatt:STeDy:10,Suchan-Bhatt-expcog-2013,STAIRS-2012} 
\end{itemize}

We envisage to publicly release the following in the course of the Rotunde initiative:

\begin{itemize}

	\item toolsets for the semantic (e.g., qualitative, activity-theoretic) grounding of perceptual narratives

	\item abstraction-driven spatio-temporal (perceptual) data visualisation capabilities to assist in analysis, and development and debugging etc 
	
	\item datasets from ongoing experimental work 

\end{itemize}

The \rotundeTxt\ initiative will enable researchers to not only utilise its deliverables, but also compare and benchmark alternate methods with respect to the scenario datasets.

\section{Sample Setup and Activity Data}

\textbf{Setup} (Fig. \ref{fig:rotunde-scenario}).\quad An example setup for the smart meeting cinematography concept consisting of a circular room structure, pan-tilt-zoom capable cameras, depth sensing equipment (e.g., Microsoft Kinect, Softkinectic Depthsense), sound sensors.

\medskip
\medskip

\begin{figure*}
  \centering
  \setlength\tabcolsep{2pt}
    \begin{tabular}{ c c c c c c } 
      \includegraphics[width=0.16\textwidth]{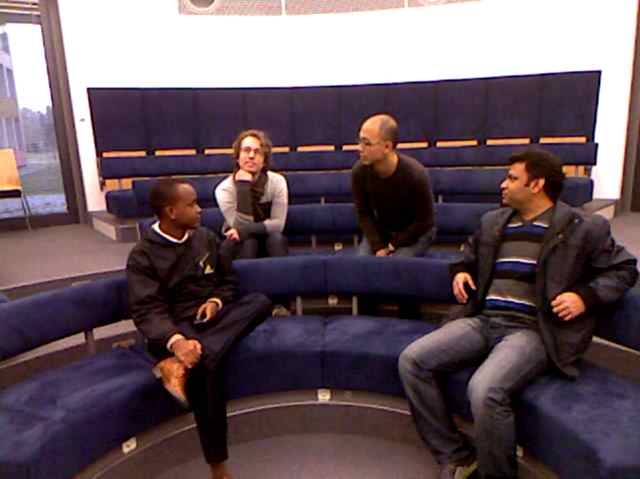} &
      \includegraphics[width=0.16\textwidth]{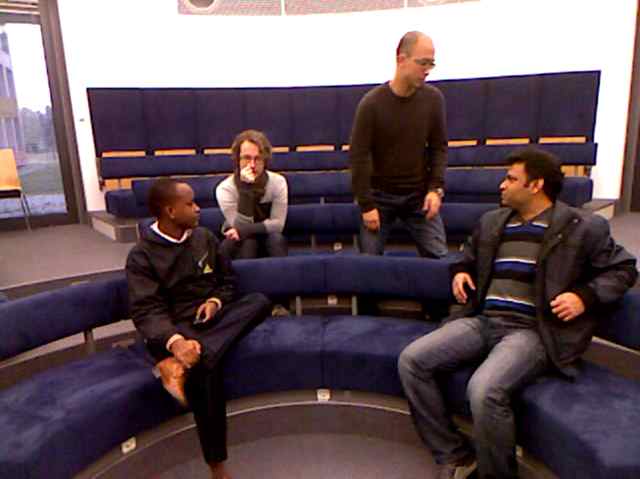} &
      \includegraphics[width=0.16\textwidth]{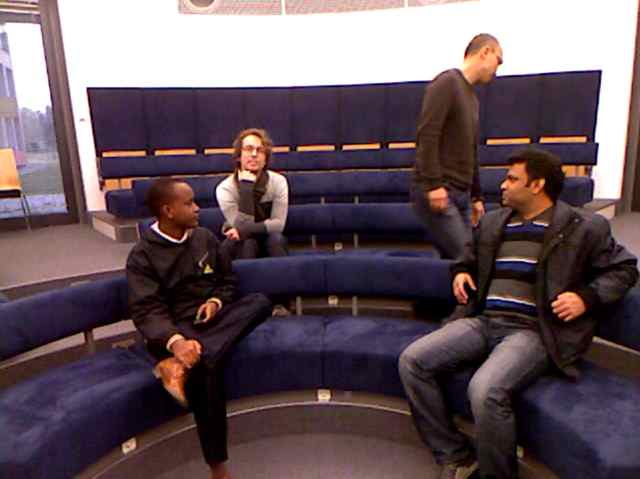} &
      \includegraphics[width=0.16\textwidth]{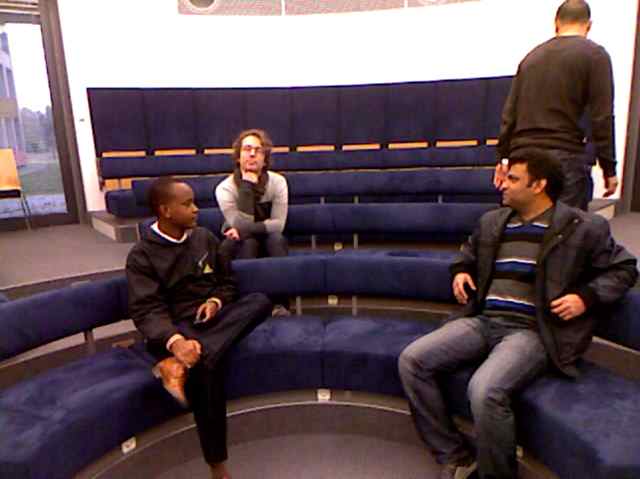} &
      \includegraphics[width=0.16\textwidth]{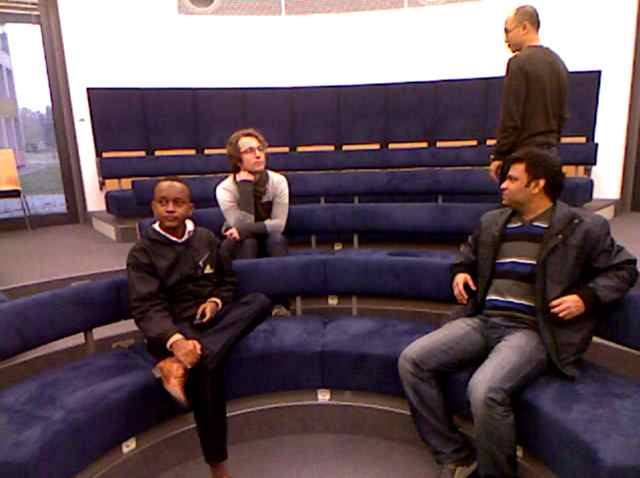} &
      \includegraphics[width=0.16\textwidth]{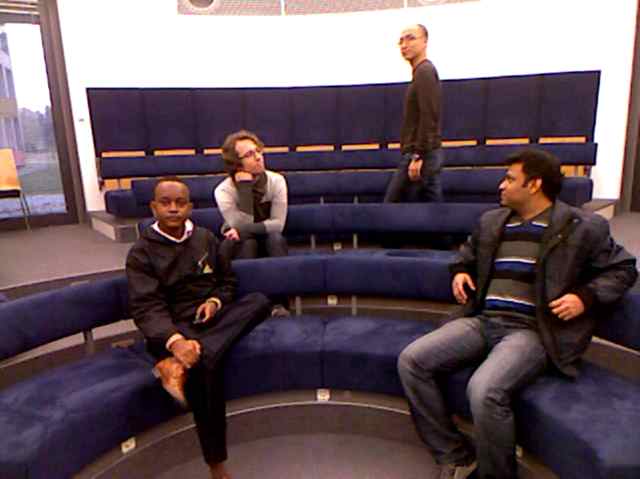} \\

      \includegraphics[width=0.16\textwidth]{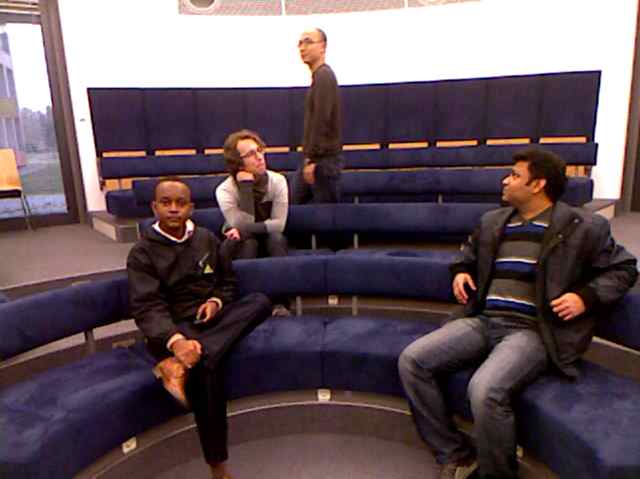} &
      \includegraphics[width=0.16\textwidth]{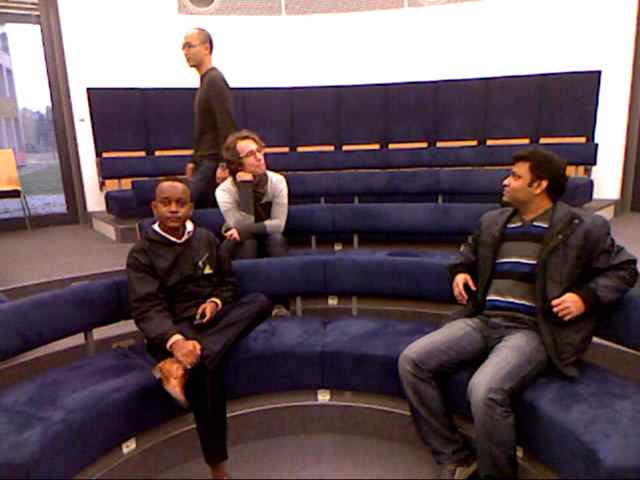} &
      \includegraphics[width=0.16\textwidth]{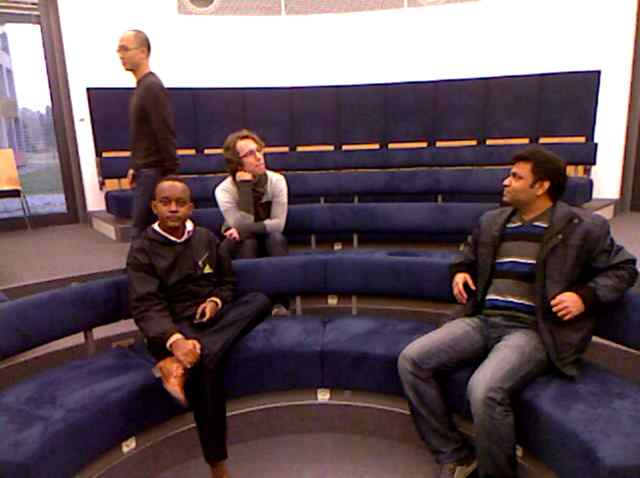} &
      \includegraphics[width=0.16\textwidth]{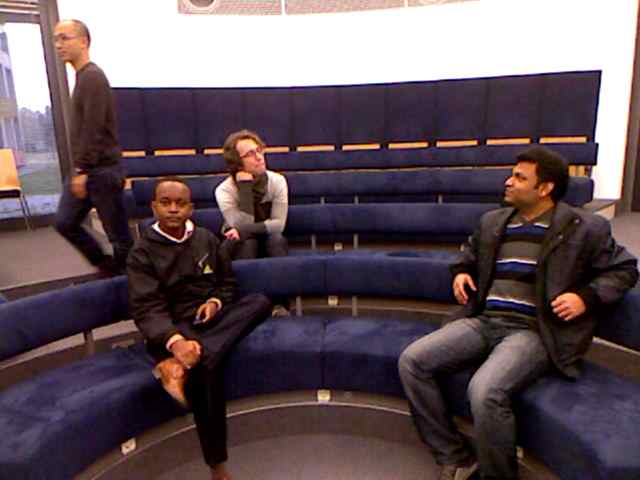} &
      \includegraphics[width=0.16\textwidth]{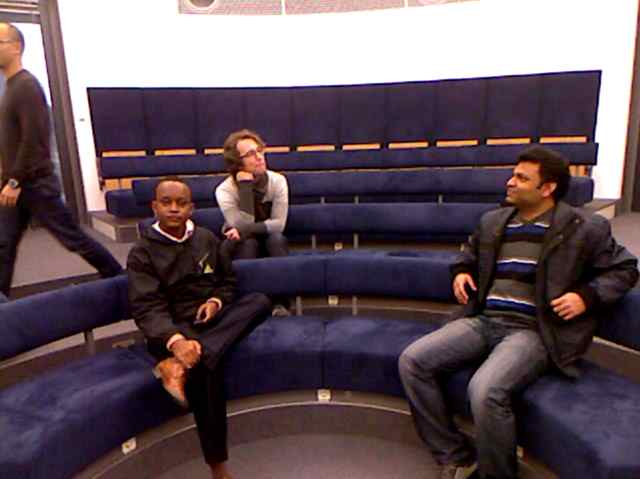} &
      \includegraphics[width=0.16\textwidth]{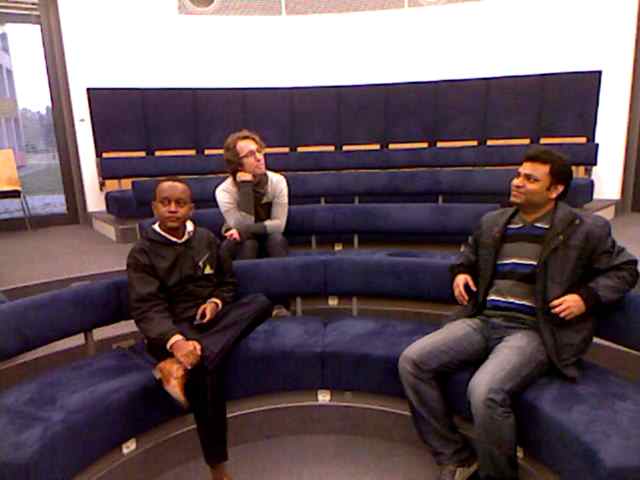} \\

&&&&&\\

      \includegraphics[width=0.16\textwidth]{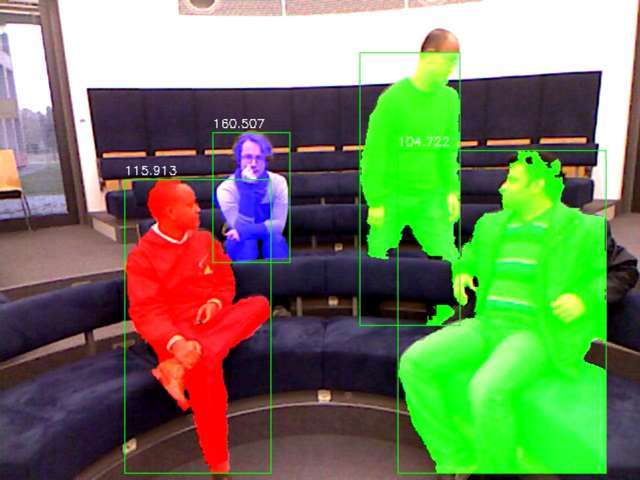} &
      \includegraphics[width=0.16\textwidth]{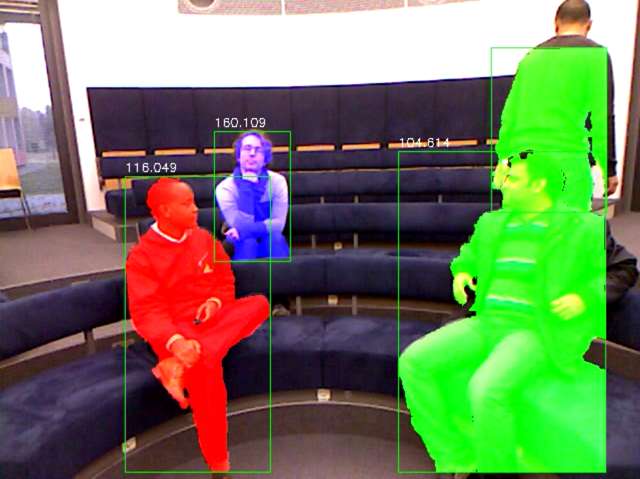} &
      \includegraphics[width=0.16\textwidth]{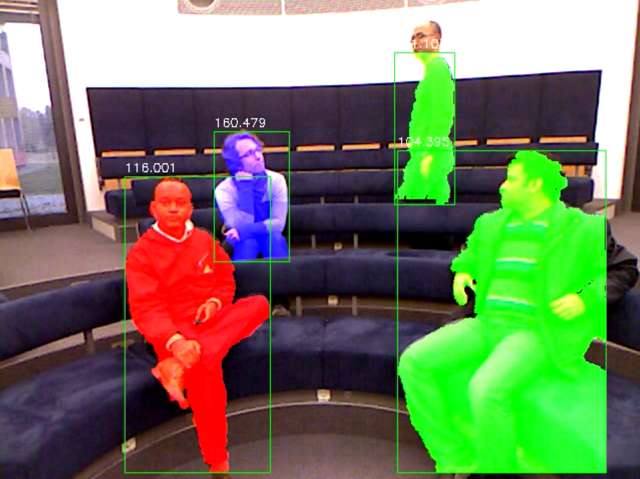} &
      \includegraphics[width=0.16\textwidth]{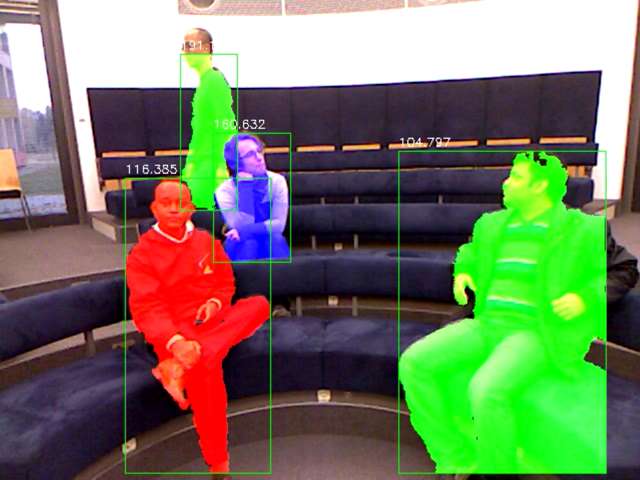} &
      \includegraphics[width=0.16\textwidth]{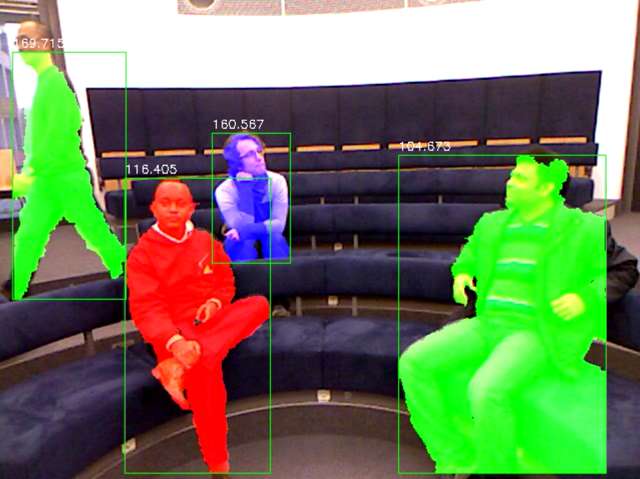} &
      \includegraphics[width=0.16\textwidth]{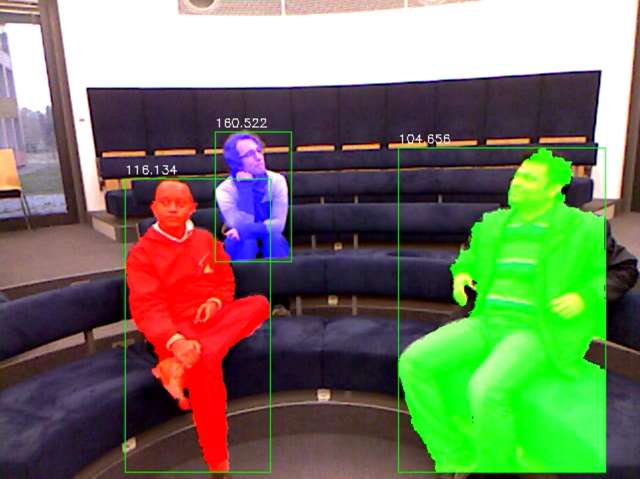} \\

&&&&&\\

      \includegraphics[width=0.16\textwidth]{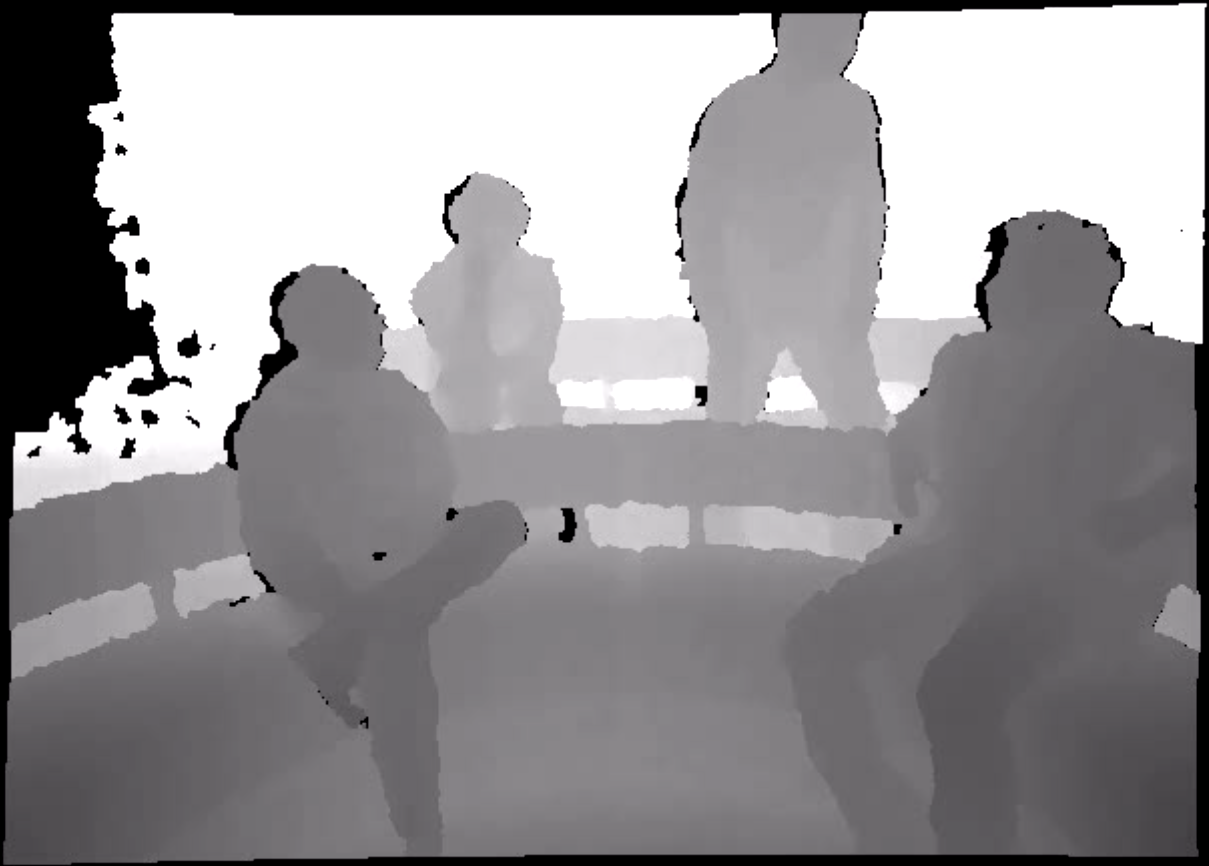} &
      \includegraphics[width=0.16\textwidth]{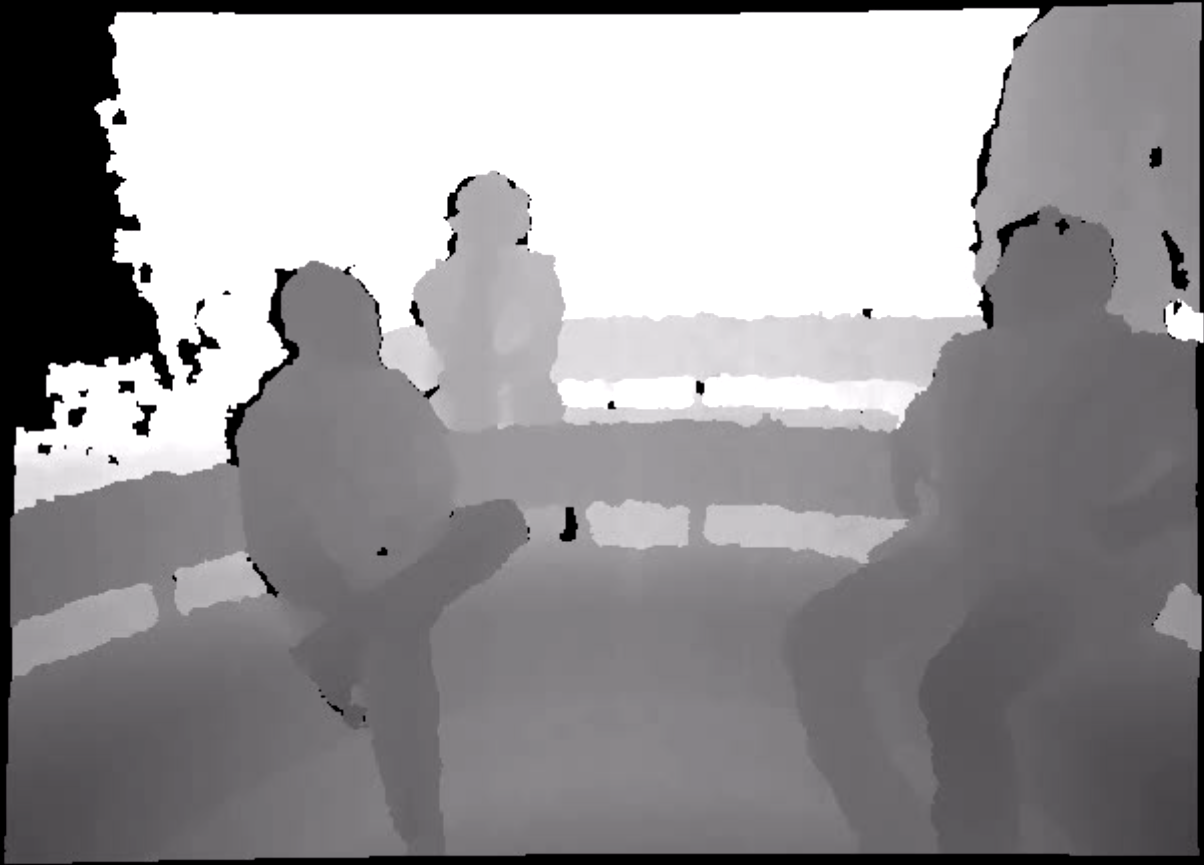} &
      \includegraphics[width=0.16\textwidth]{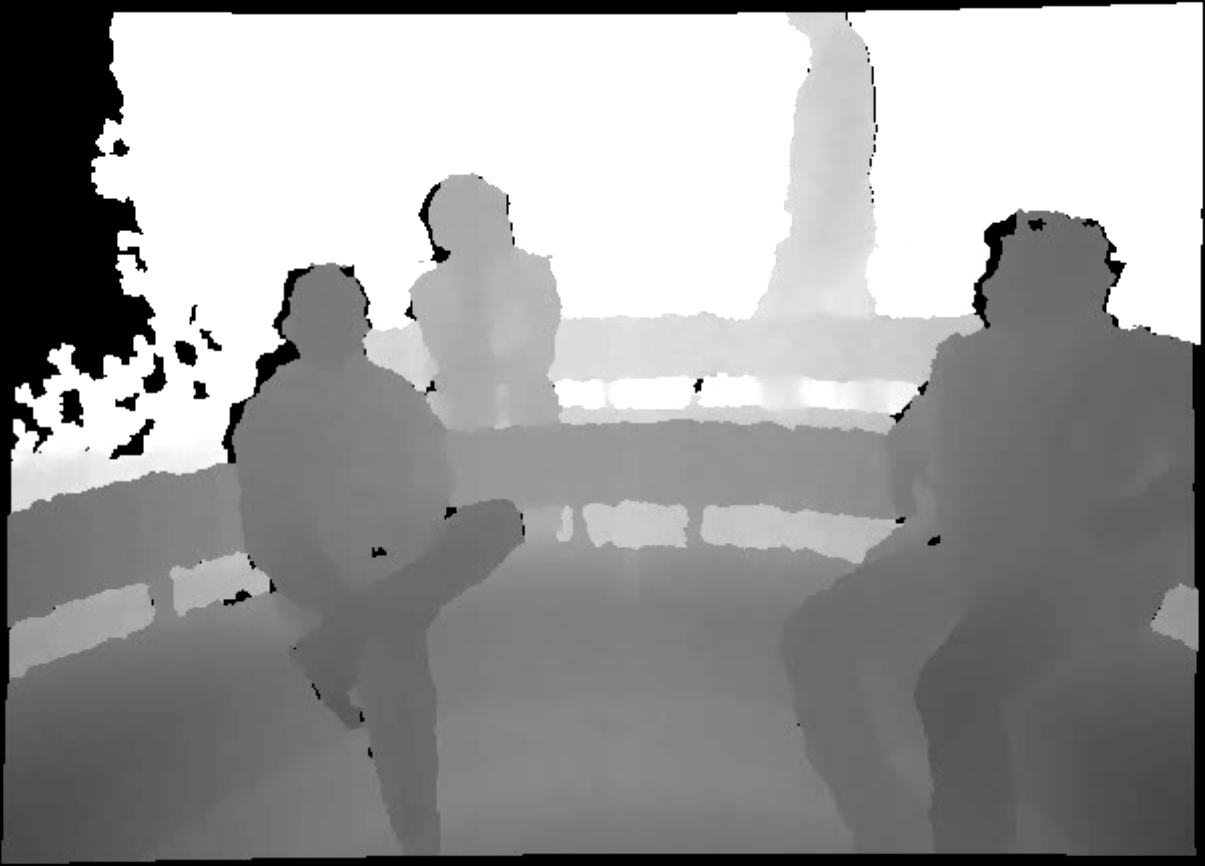} &
      \includegraphics[width=0.16\textwidth]{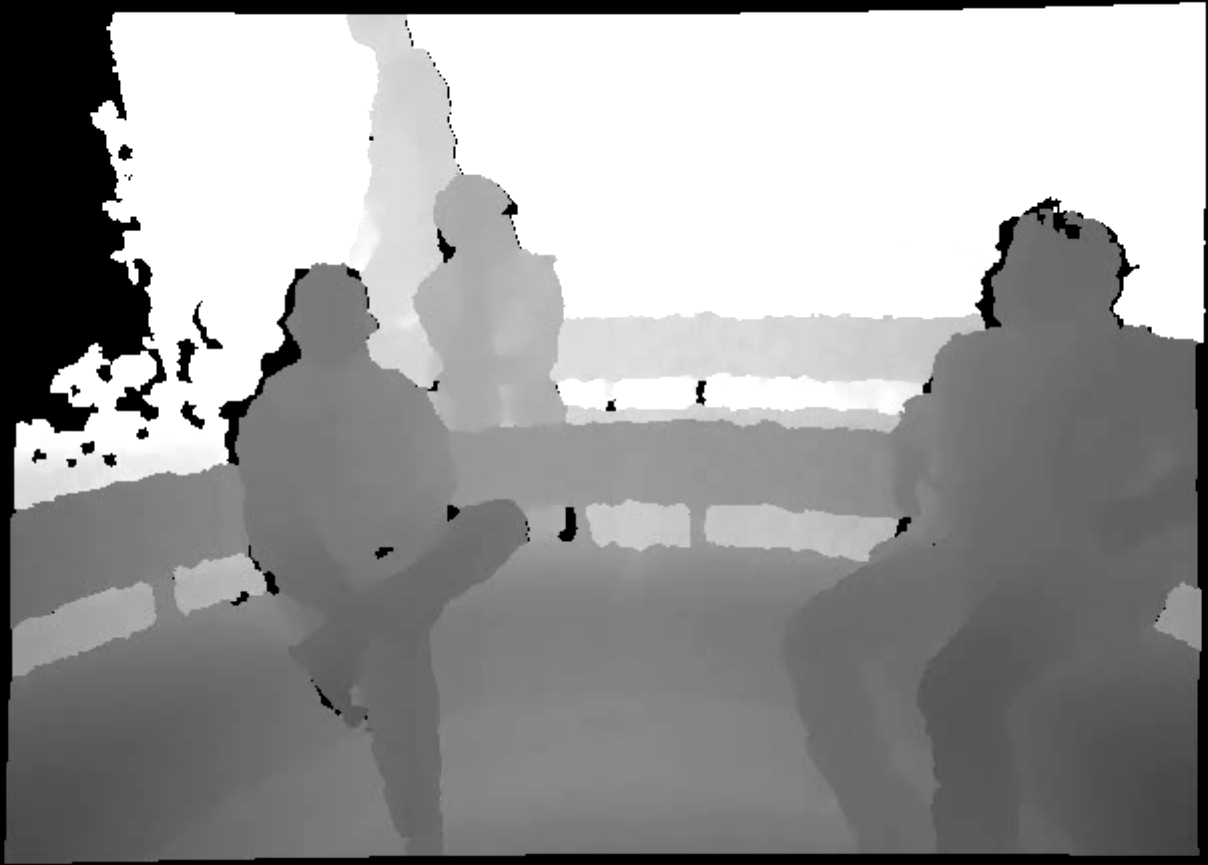} &
      \includegraphics[width=0.16\textwidth]{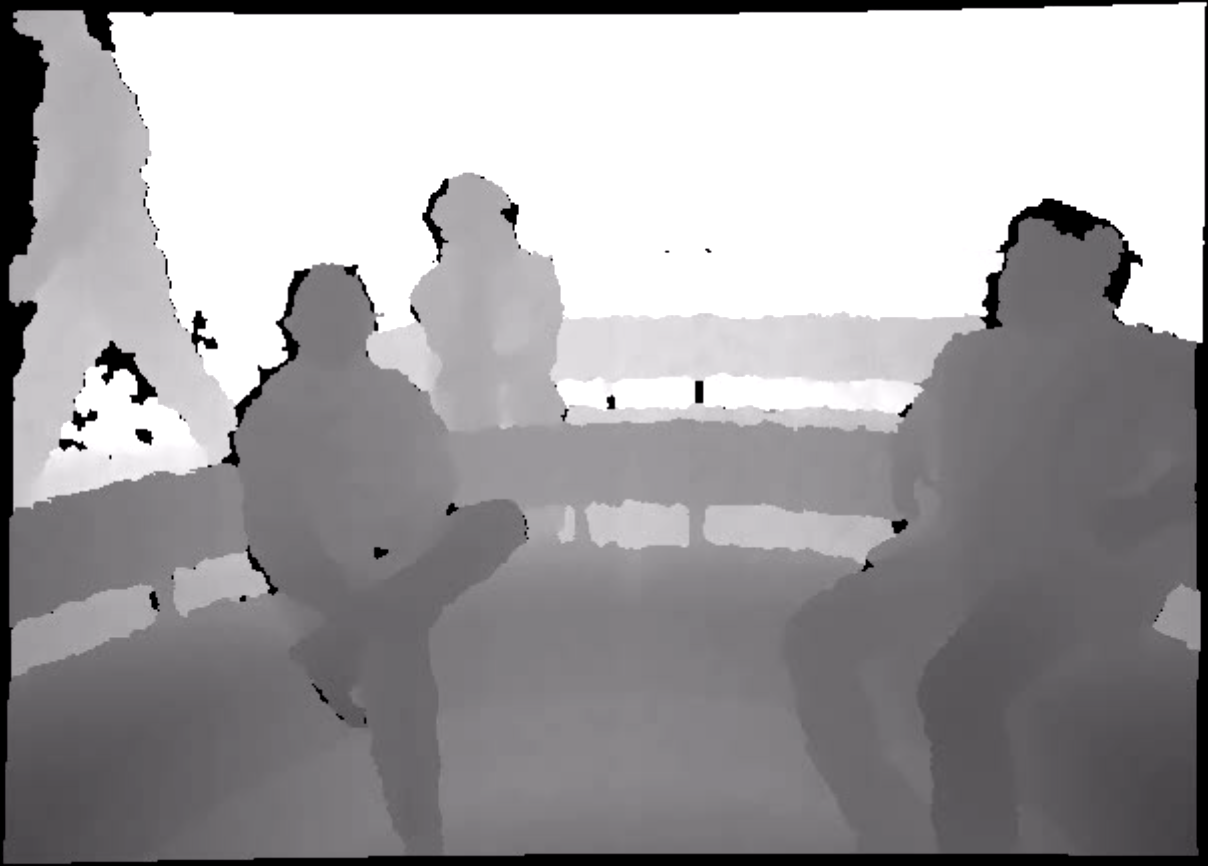} &
      \includegraphics[width=0.16\textwidth]{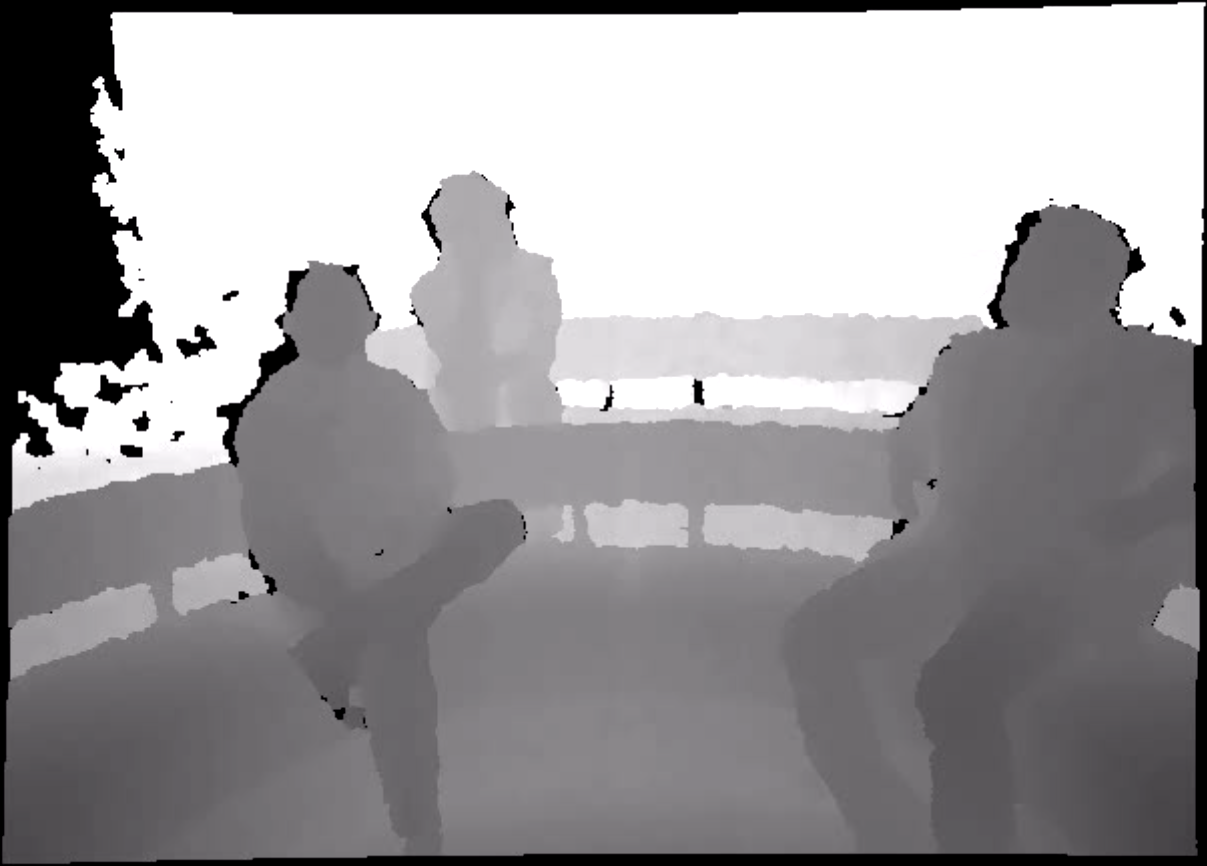} \\

&&&&&\\

      \multicolumn{6}{c }{\includegraphics[width=1\textwidth]{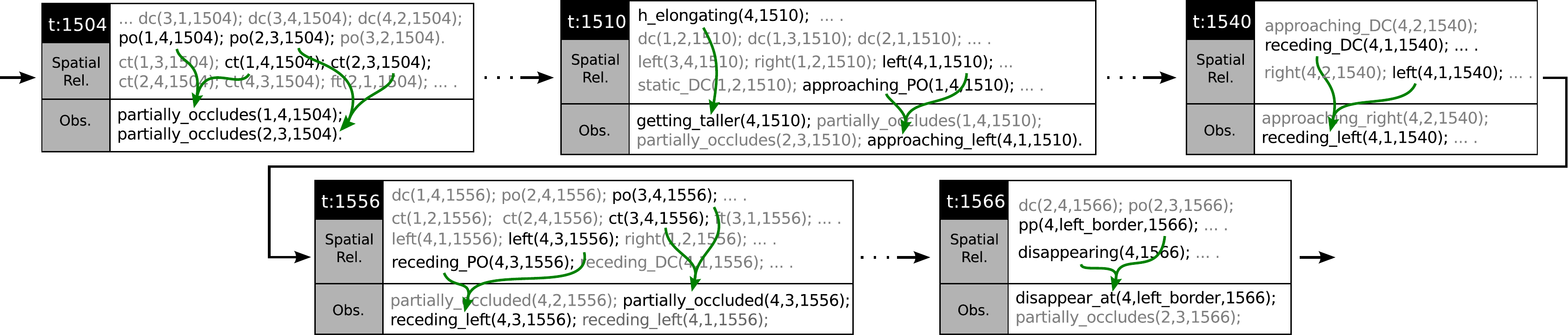}}
    \end{tabular}

    \caption{Activity Sequence: \emph{leave meeting}, corresponding RGB and Depth data, and high-level declarative models}

  \label{fig:activity-leave-meeting}
\end{figure*}

\begin{figure*}
  \centering
  \setlength\tabcolsep{2pt}
    \begin{tabular}{ c c c c c c } 
      \includegraphics[width=0.16\textwidth]{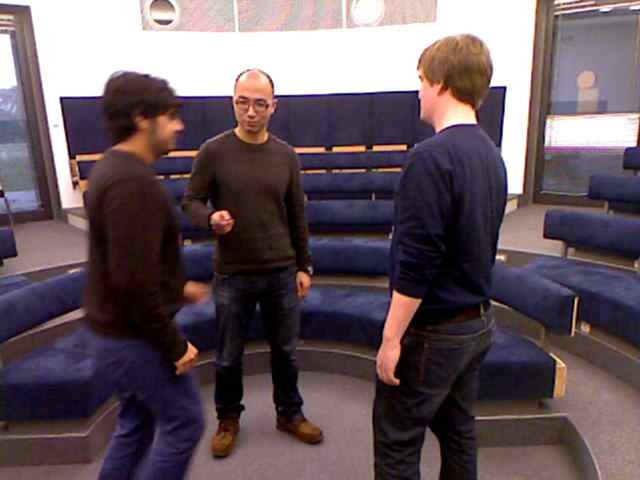} &
      \includegraphics[width=0.16\textwidth]{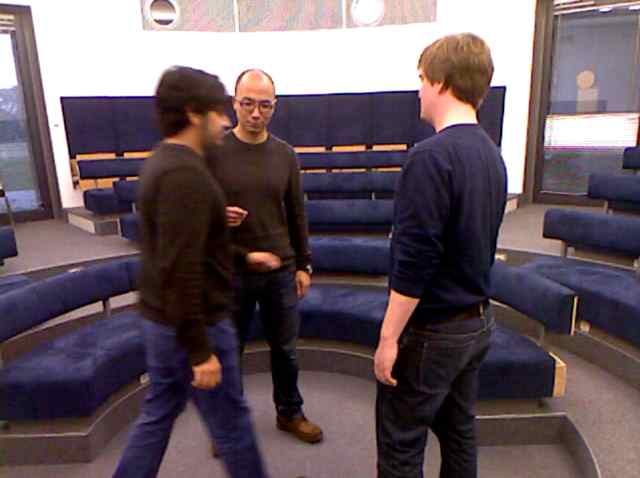} &
      \includegraphics[width=0.16\textwidth]{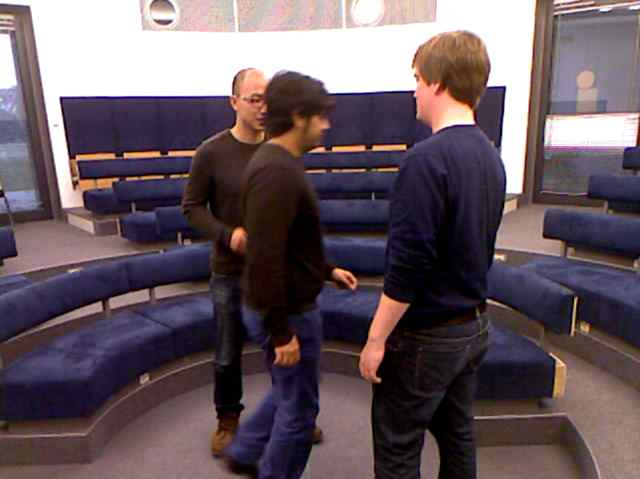} &
      \includegraphics[width=0.16\textwidth]{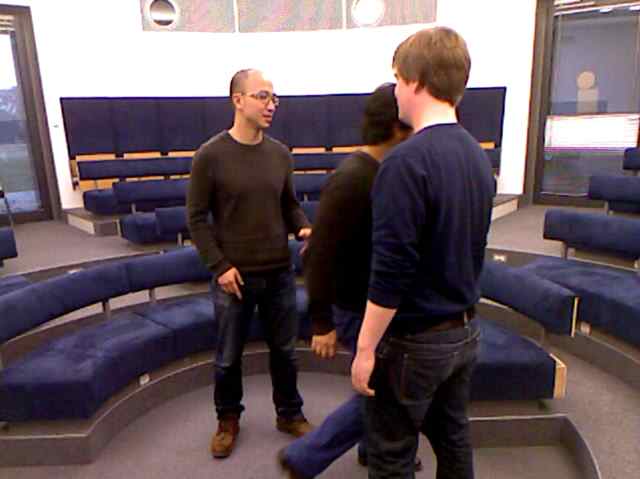} &
      \includegraphics[width=0.16\textwidth]{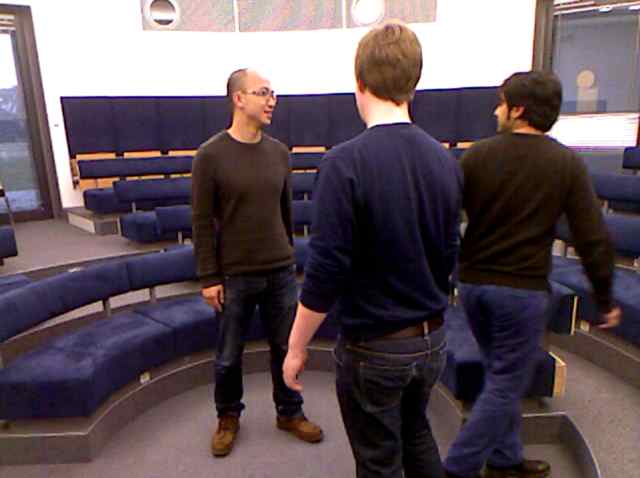} &
      \includegraphics[width=0.16\textwidth]{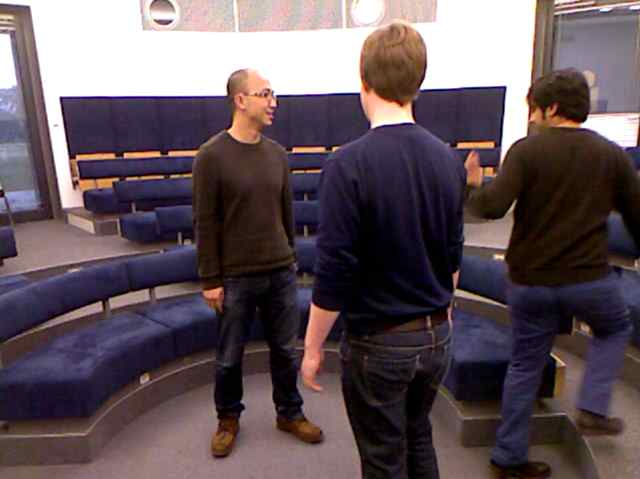} \\

&&&&&\\

      \includegraphics[width=0.16\textwidth]{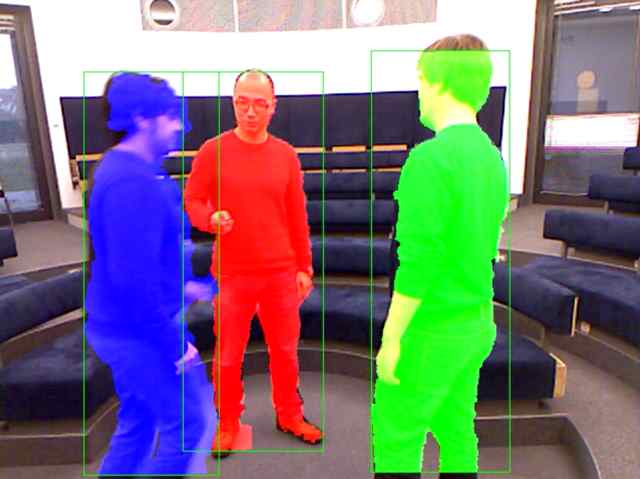} &
      \includegraphics[width=0.16\textwidth]{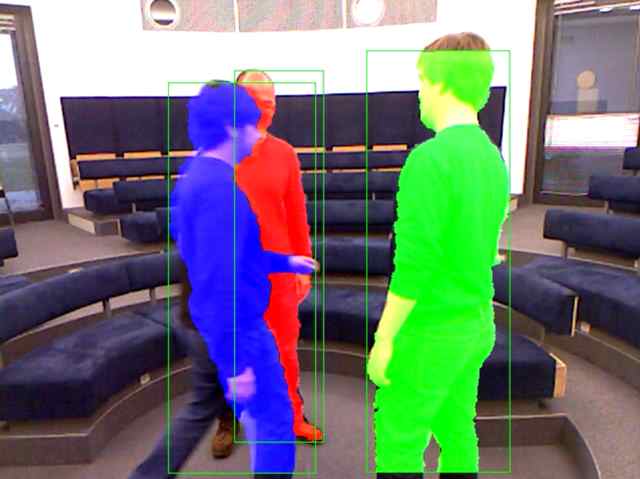} &
      \includegraphics[width=0.16\textwidth]{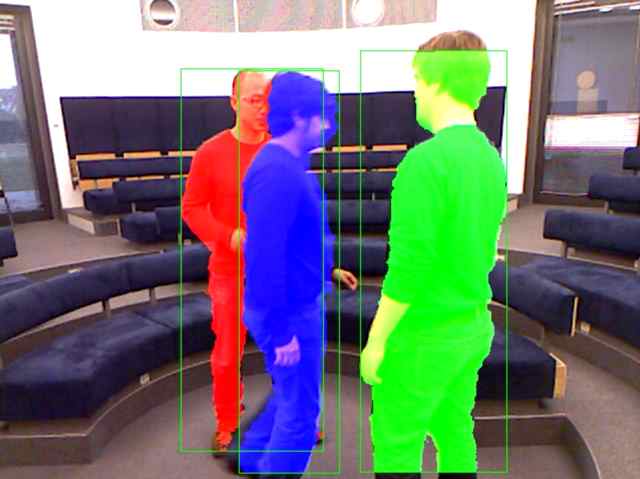} &
      \includegraphics[width=0.16\textwidth]{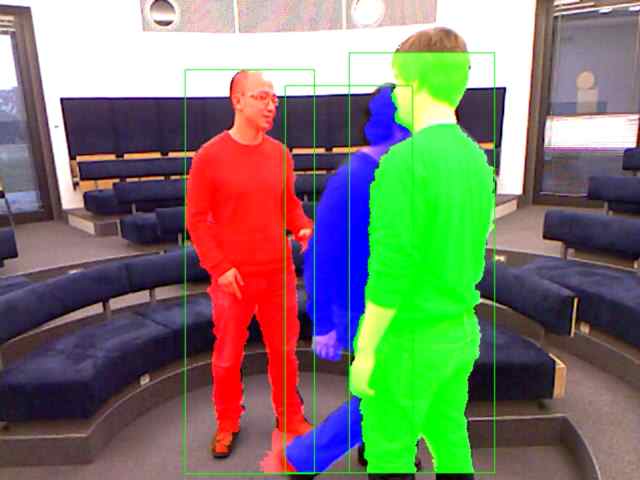} &
      \includegraphics[width=0.16\textwidth]{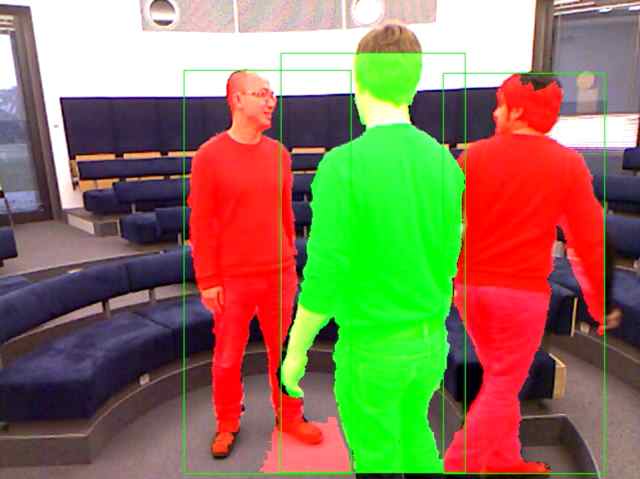} &
      \includegraphics[width=0.16\textwidth]{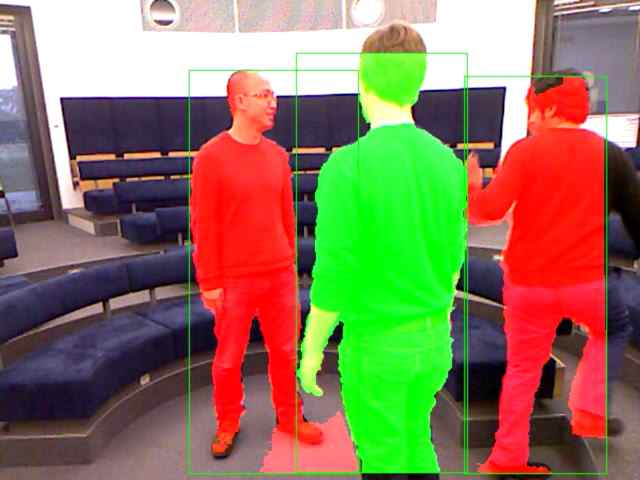} \\

&&&&&\\

      \includegraphics[width=0.16\textwidth]{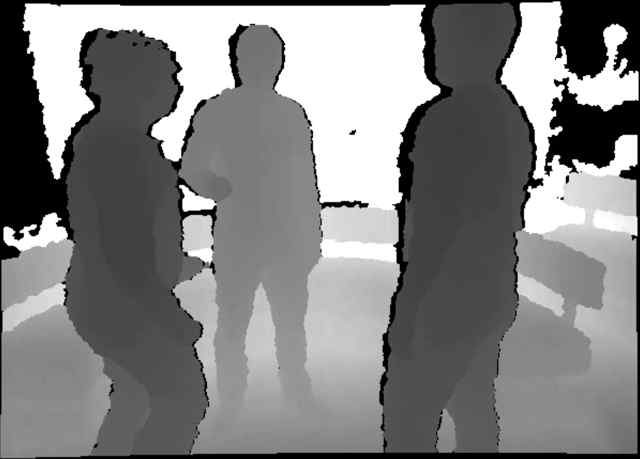} &
      \includegraphics[width=0.16\textwidth]{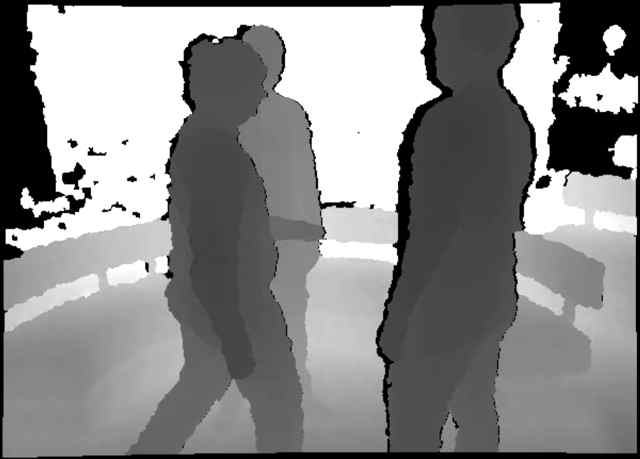} &
      \includegraphics[width=0.16\textwidth]{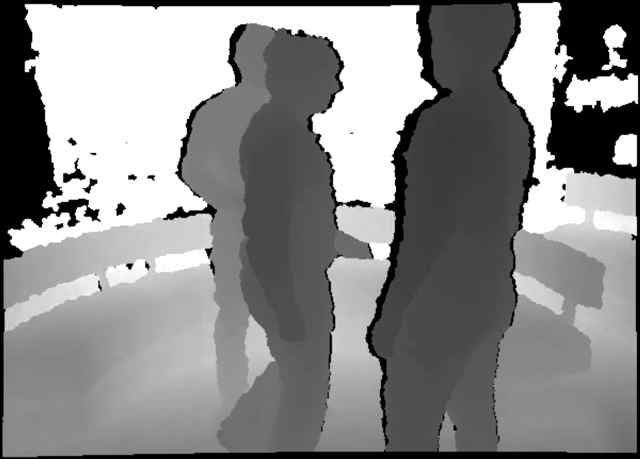} &
      \includegraphics[width=0.16\textwidth]{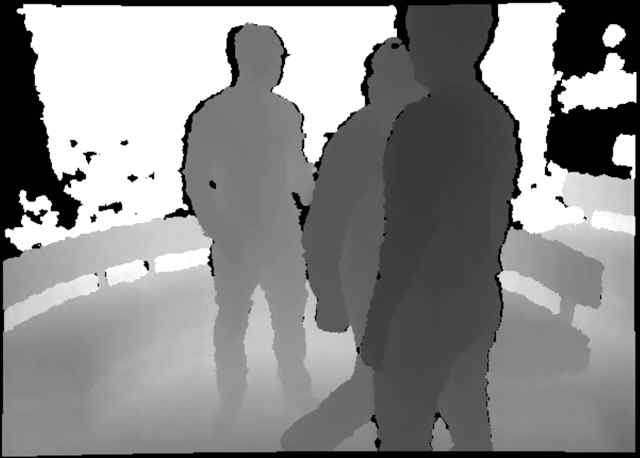} &
      \includegraphics[width=0.16\textwidth]{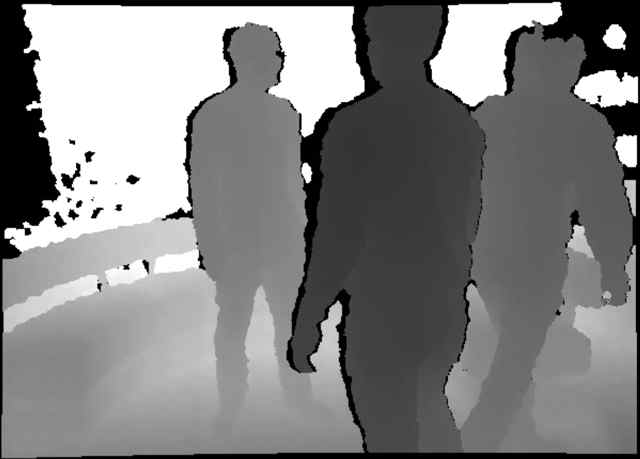} &
      \includegraphics[width=0.16\textwidth]{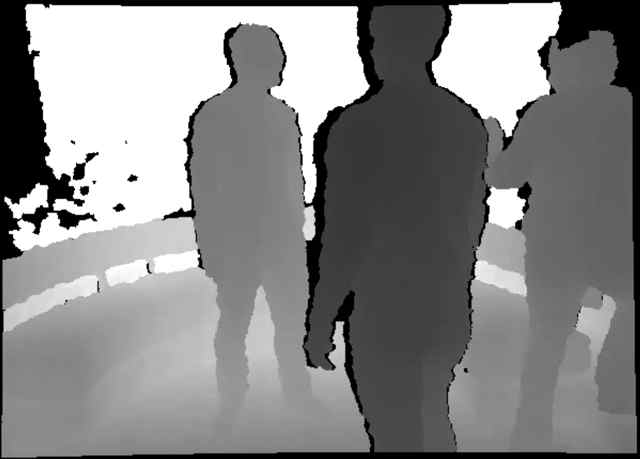} \\
    \end{tabular}
      \caption{Activity Sequence: \emph{passing in-between people}, corresponding RGB and Depth profile data}   
  \label{fig:activity-passing-through}
\end{figure*}

\begin{figure*}
  \centering
  \setlength\tabcolsep{2pt}
    \begin{tabular}{ c c c c } 
      \includegraphics[width=0.24\textwidth]{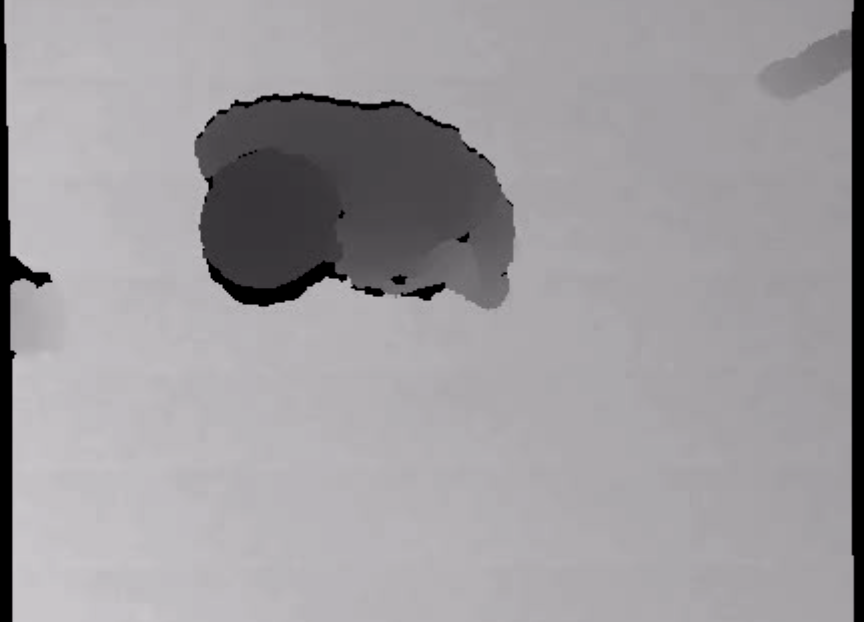} &
      \includegraphics[width=0.24\textwidth]{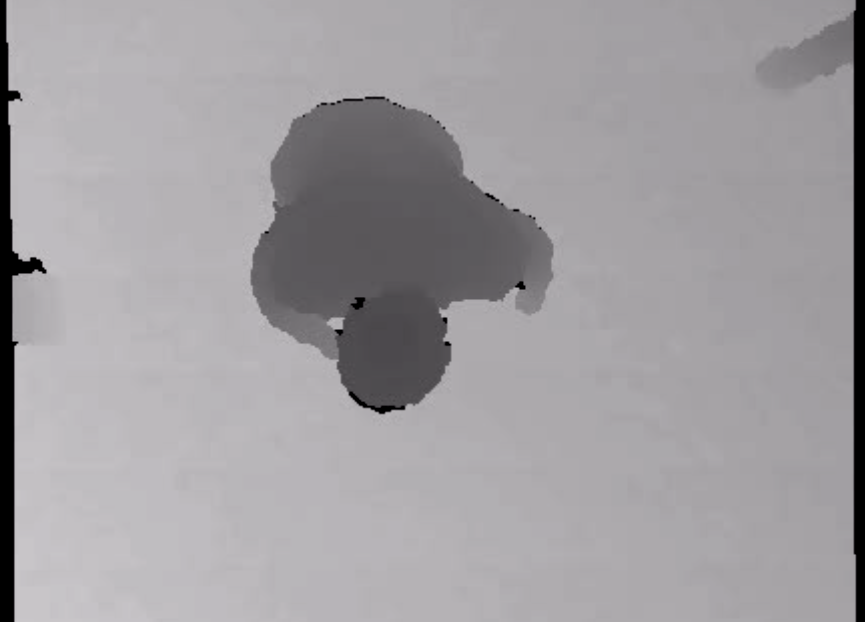} &
      \includegraphics[width=0.24\textwidth]{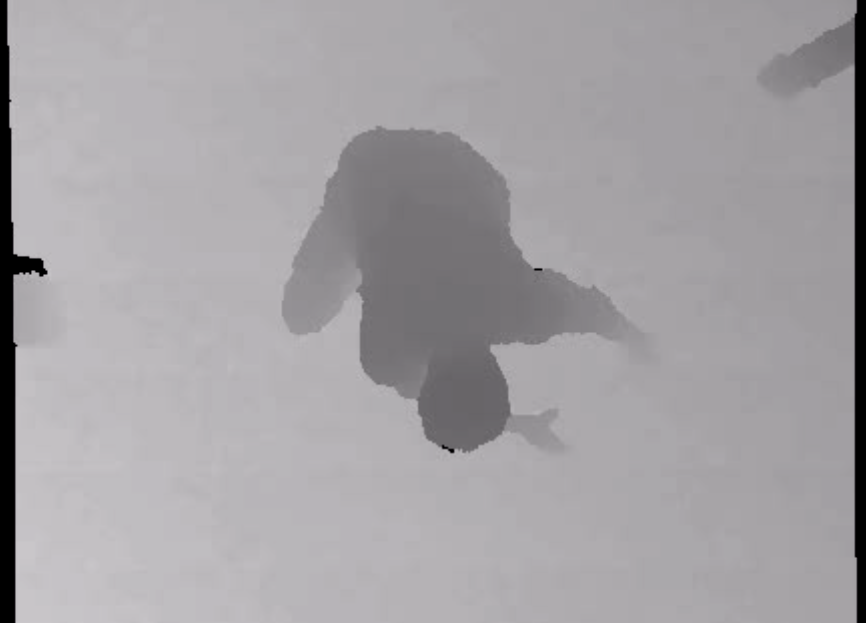} &
      \includegraphics[width=0.24\textwidth]{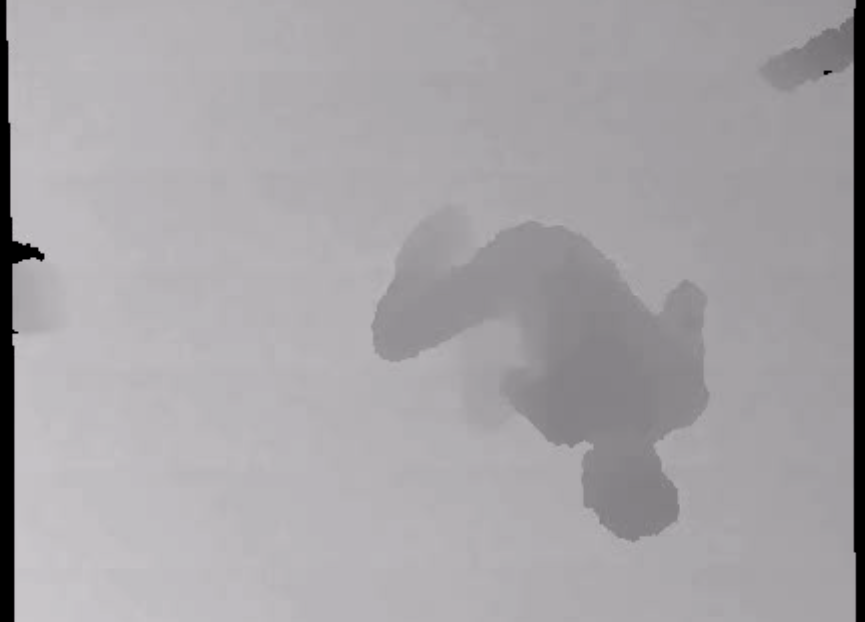} \\

&&&\\

      \includegraphics[width=0.24\textwidth]{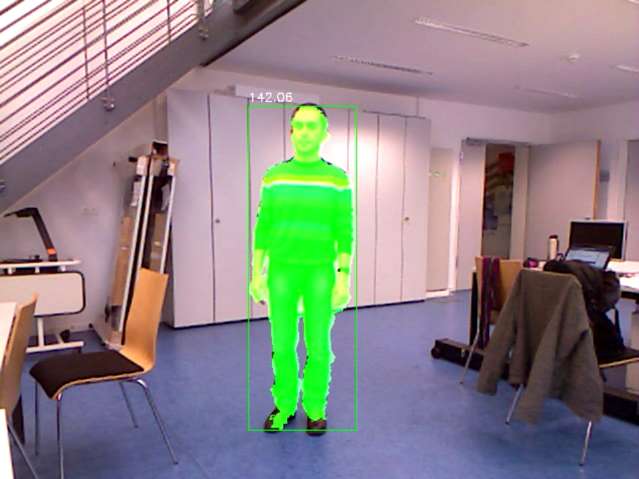} &
      \includegraphics[width=0.24\textwidth]{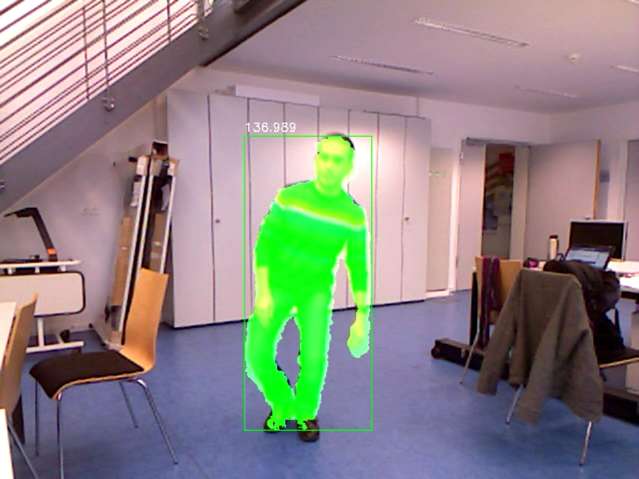} &
      \includegraphics[width=0.24\textwidth]{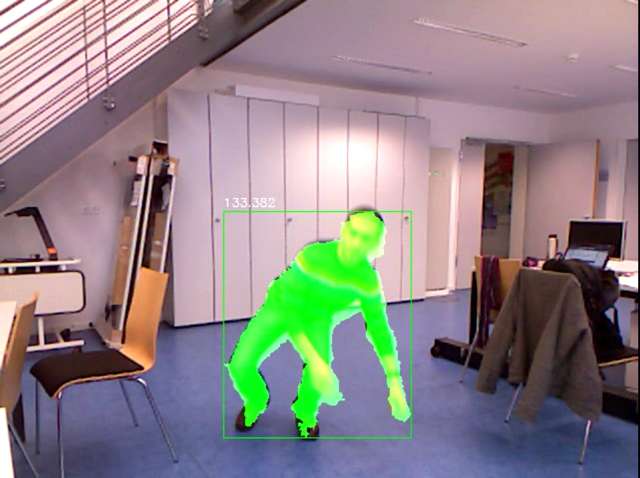} &
      \includegraphics[width=0.24\textwidth]{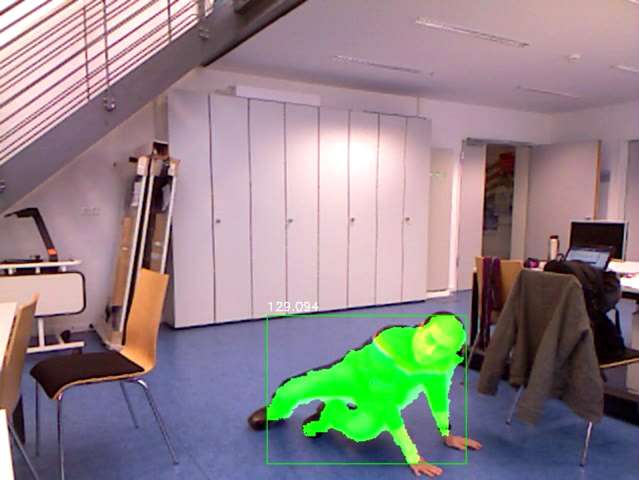} \\

&&&\\

      \includegraphics[width=0.24\textwidth]{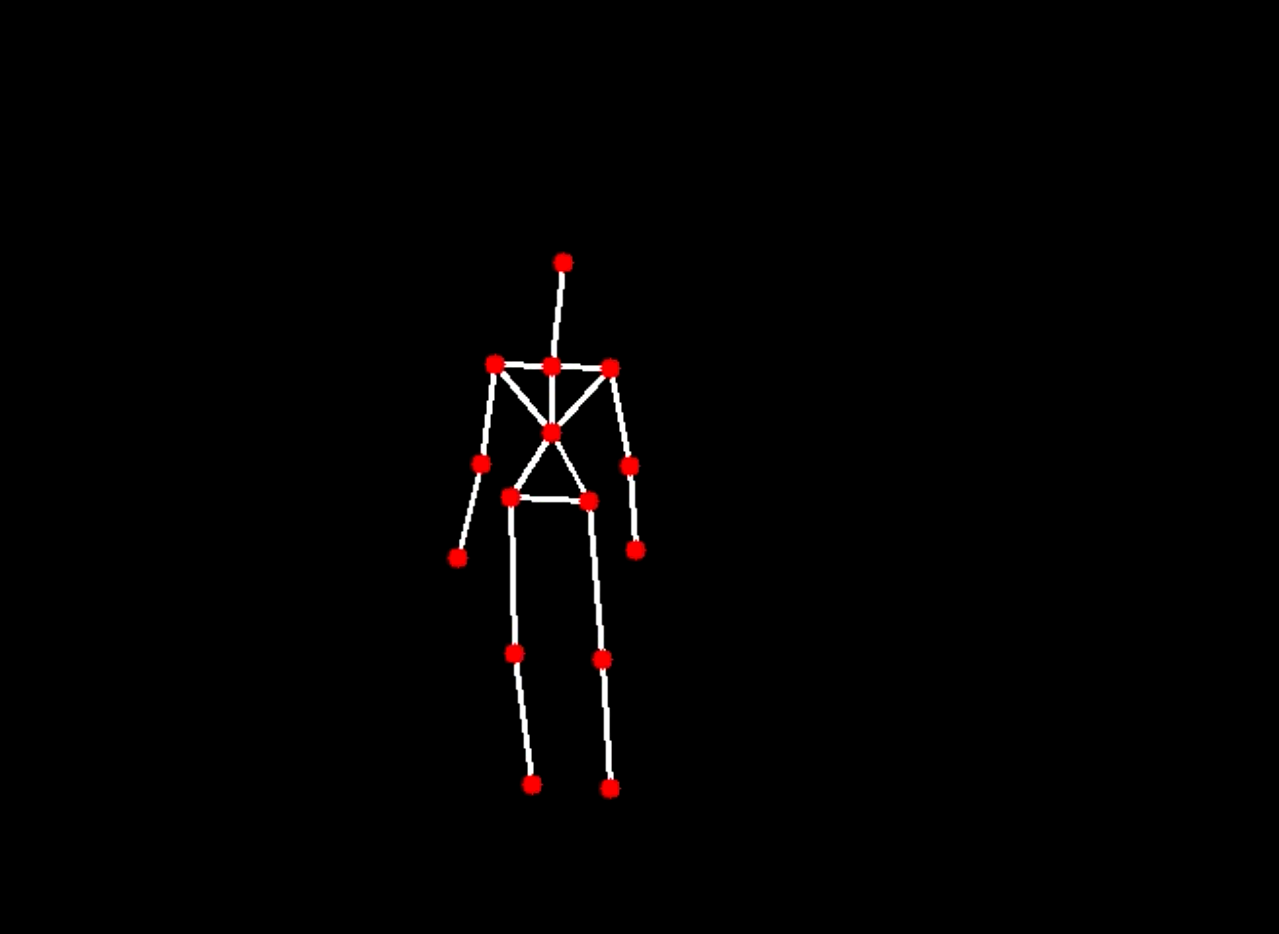} &
      \includegraphics[width=0.24\textwidth]{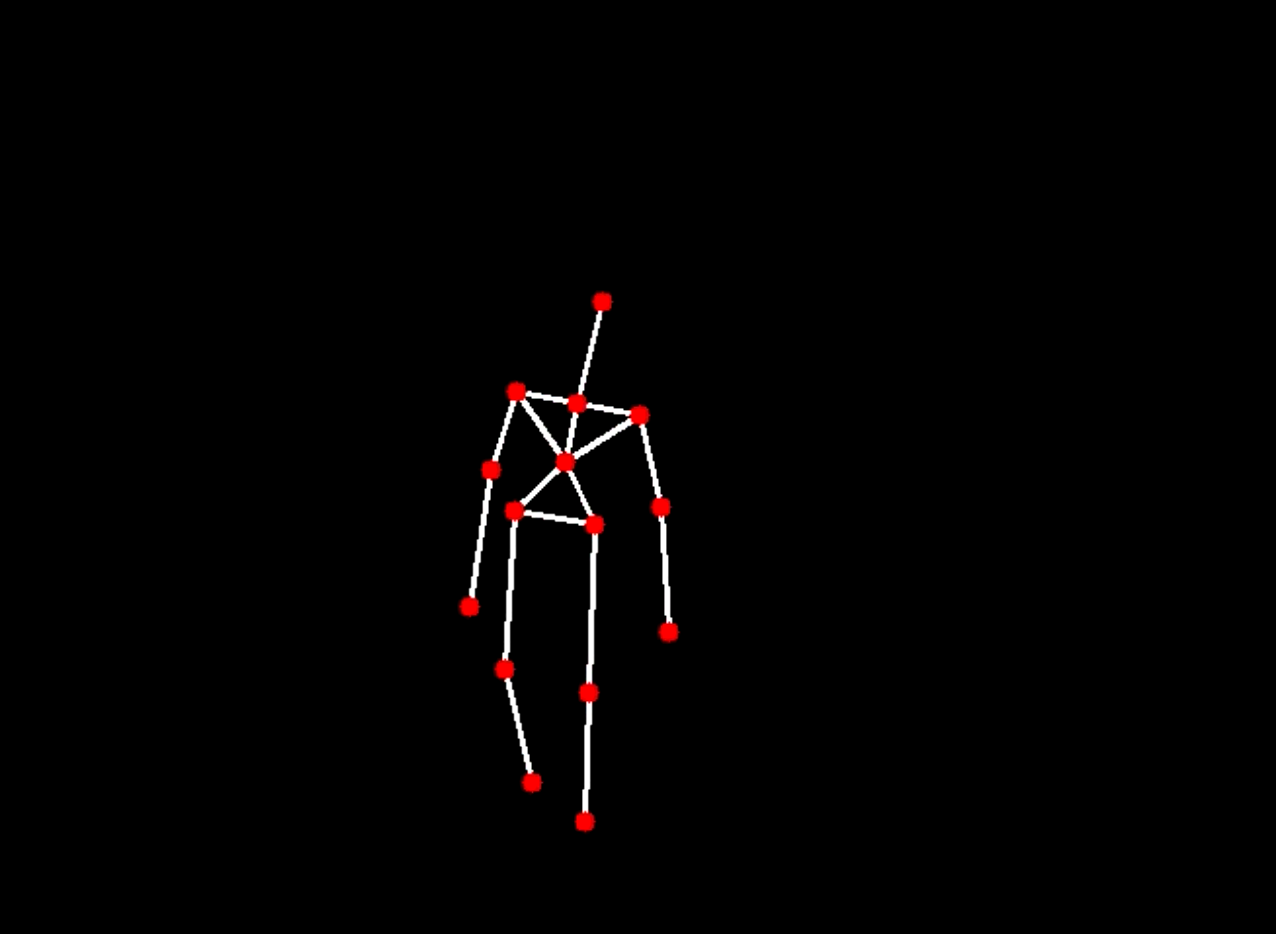} &
      \includegraphics[width=0.24\textwidth]{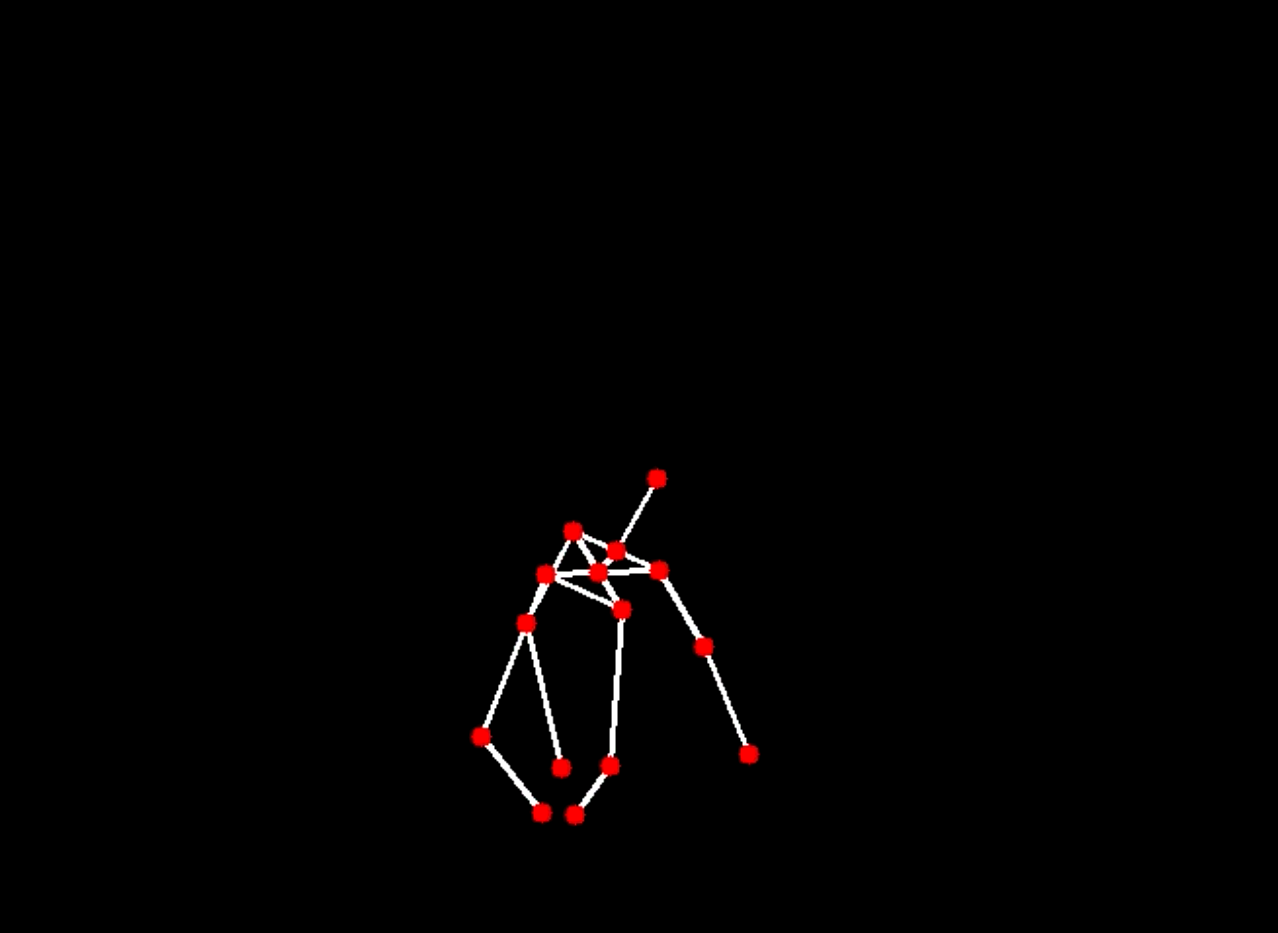} &
      \includegraphics[width=0.24\textwidth]{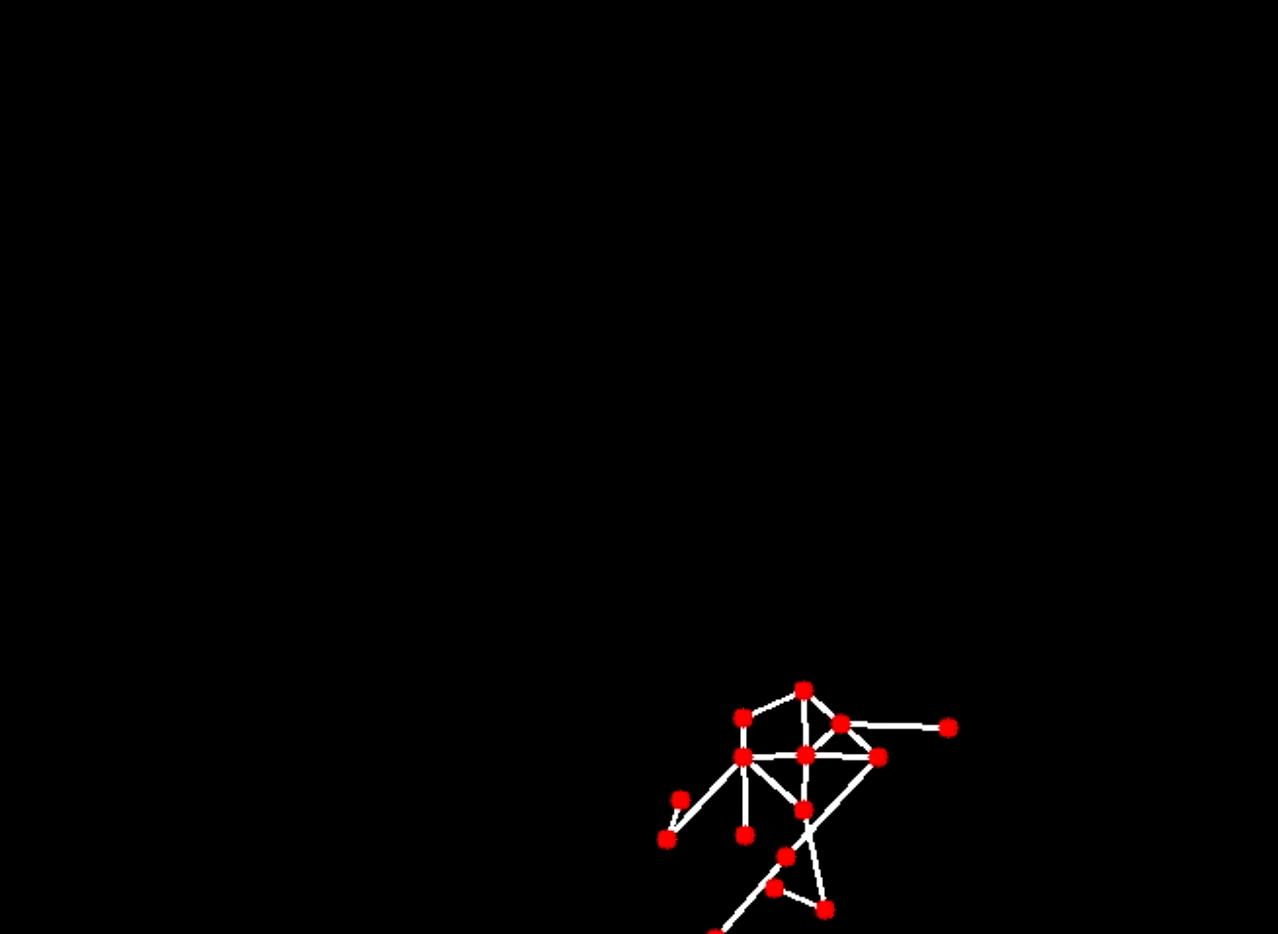} \\
    \end{tabular}
     \caption{Activity Sequence: \emph{falling down}, corresponding RGB and Depth profile data, and body-joint skeleton model}   
  \label{fig:activity-falling-man}
\end{figure*}

\textbf{Activity Data} (Fig. \ref{fig:activity-leave-meeting}-\ref{fig:activity-falling-man}).\quad Sample scenarios and datasets consisting of:  RGB and depth profile, body skeleton data, and high-level declarative models generated from raw data for further analysis (e.g., for reasoning, learning, control).

\medskip

Activity Sequence:\quad \emph{\textcolor{black}{leave meeting}}, corresponding RGB and Depth data, and high-level declarative models (Fig. \ref{fig:activity-leave-meeting})

\medskip

Activity Sequence:\quad \emph{passing in-between people}, corresponding RGB and Depth profile data (Fig. \ref{fig:activity-passing-through})

\medskip

Activity Sequence:\quad \emph{falling down}, corresponding RGB and Depth profile data, and body-joint skeleton model (Fig. \ref{fig:activity-falling-man})

\section{Acknowledgements}
{\small The preliminary concept for the Rotunde initiative and its developmental and benchmarking agenda were presented at the Dagstuhl Seminars ``12491 -- \emph{Interpreting Observed Action}'' (S. Biundo-Stephan, H. W. Guesgen, J. Hertzberg., and S. Marsland); and ``12492 -- \emph{Human Activity Recognition in Smart Environments} (J. Begole, J. Crowley, P. Lukowicz, A. Schmidt)''. We thank the seminar participants for discussions, feedback, and impulses.

We gratefully acknowledge funding by the DFG Spatial Cognition Research Center (SFB/TR 8).
}

{

}


\begin{thebibliography}{}

\bibitem[\protect\citeauthoryear{Bhatt and Flanagan}{2010}]{Bhatt:STeDy:10}
Bhatt, M., and Flanagan, G.
\newblock 2010.
\newblock {Spatio-Temporal Abduction for Scenario and Narrative Completion}.
\newblock In {\em Proceedings of the International Workshop on Spatio-Temporal
  Dynamics, co-located with the European Conference on Artificial Intelligence
  (ECAI-10)},  31--36.
\newblock ECAI Workshop Proceedings., and SFB/TR 8 Spatial Cognition Report
  Series.

\bibitem[\protect\citeauthoryear{Bhatt, Lee, and
  Schultz}{2011}]{bhatt-et-al-2011}
Bhatt, M.; Lee, J.~H.; and Schultz, C.
\newblock 2011.
\newblock {CLP(QS): A Declarative Spatial Reasoning Framework}.
\newblock In {\em COSIT: Conference on Spatial Information Theory},  210--230.

\bibitem[\protect\citeauthoryear{Bhatt, Suchan, and
  Schultz}{2013}]{Bhatt2013-CMN}
Bhatt, M.; Suchan, J.; and Schultz, C.
\newblock 2013.
\newblock {Cognitive Interpretation of Everyday Activities -- Toward Perceptual
  Narrative Based Visuo-Spatial Scene Interpretation}.
\newblock In Finlayson, M.; Fisseni, B.; Loewe, B.; and Meister, J.~C., eds.,
  {\em Computational Models of Narrative (CMN) 2013., a satellite workshop of
  CogSci 2013: The 35th meeting of the Cognitive Science Society.}

\bibitem[\protect\citeauthoryear{Bhatt}{2012}]{Bhatt:RSAC:2012}
Bhatt, M.
\newblock 2012.
\newblock Reasoning about space, actions and change: A paradigm for
  applications of spatial reasoning.
\newblock In {\em Qualitative Spatial Representation and Reasoning: Trends and
  Future Directions}.
\newblock IGI Global, USA.

\bibitem[\protect\citeauthoryear{Dubba \bgroup et al\mbox.\egroup
  }{2012}]{dubba-bhatt-2012}
Dubba, K.; Bhatt, M.; Dylla, F.; Hogg, D.; and Cohn, A.
\newblock 2012.
\newblock Interleaved inductive-abductive reasoning for learning complex event
  models.
\newblock In Muggleton, S.; Tamaddoni-Nezhad, A.; and Lisi, F., eds., {\em
  Inductive Logic Programming}, volume 7207 of {\em Lecture Notes in Computer
  Science}. Springer Berlin / Heidelberg.
\newblock  113--129.

\bibitem[\protect\citeauthoryear{Eppe and Bhatt}{2013}]{CR-2013-Narra-CogRob}
Eppe, M., and Bhatt, M.
\newblock 2013.
\newblock {Narrative based Postdictive Reasoning for Cognitive Robotics}.
\newblock In {\em COMMONSENSE 2013: 11th International Symposium on Logical
  Formalizations of Commonsense Reasoning}.
\newblock (to appear).

\bibitem[\protect\citeauthoryear{Suchan and Bhatt}{2012}]{STAIRS-2012}
Suchan, J., and Bhatt, M.
\newblock 2012.
\newblock Toward an activity theory based model of spatio-temporal interactions
  - integrating situational inference and dynamic (sensor) control.
\newblock In Kersting, K., and Toussaint, M., eds., {\em STAIRS}, volume 241 of
  {\em Frontiers in Artificial Intelligence and Applications},  318--329.
\newblock IOS Press.

\bibitem[\protect\citeauthoryear{Suchan and
  Bhatt}{2013}]{Suchan-Bhatt-expcog-2013}
Suchan, J., and Bhatt, M.
\newblock 2013.
\newblock {The ExpCog Framework: High-Level Spatial Control and Planning for
  Cognitive Robotics}.
\newblock In {\em Bridges between the Methodological and Practical Work of the
  Robotics and Cognitive Systems Communities - From Sensors to Concepts}.
\newblock Intelligent Systems Reference Library, Springer.
\newblock (in press).

\end{thebibliography}
\end{document}